\documentclass[acmsmall]{acmart}
\usepackage{xcolor}
\usepackage{subfig}
\usepackage{enumitem}
\usepackage[ruled]{algorithm2e}
\usepackage{algorithmic}

\setcopyright{licensedothergov}
\acmYear{2023} 
\acmVolume{6} 
\acmNumber{2} 
\acmArticle{1} 
\acmMonth{8} 
\acmPrice{15.00}
\acmDOI{10.1145/3606926}

\begin{document}

\title{MAAIP: Multi-Agent Adversarial Interaction Priors for imitation from fighting demonstrations for physics-based characters}

\author{Mohamed Younes}
\orcid{0000-0002-4831-3343}
\affiliation{
  \institution{Inria, IRISA, University of Rennes}
  \city{Rennes}
  \country{France}
}
\email{mohamed.younes@inria.fr}

\author{Ewa Kijak}
\orcid{0000-0003-0232-1097}
\affiliation{
\institution{University of Rennes, Inria, IRISA}
\city{Rennes}
\country{France}
}
\email{ewa.kijak@irisa.fr}

\author{Richard Kulpa}
\orcid{0000-0002-1863-8921}
\affiliation{
\institution{University Rennes 2, Inria, M2S}
\city{Rennes}
\country{France}
}
\email{richard.kulpa@irisa.fr}

\author{Simon Malinowski}
\orcid{0000-0002-9663-562X}
\affiliation{
\institution{University of Rennes, Inria, IRISA}
\city{Rennes}
\country{France}
}
\email{simon.malinowski@irisa.fr}

\author{Franck Multon}
\orcid{0000-0003-2690-0077}
\affiliation{
\institution{University of Rennes, Inria, IRISA, M2S}
\city{Rennes}
\country{France}}
\email{franck.multon@irisa.fr}

\renewcommand{\shorttitle}{Multi-Agent Adversarial Interaction Priors for imitation from fighting demonstrations for physics-based characters}

\begin{abstract}
 Simulating realistic interaction and motions for physics-based characters is of great interest for interactive applications, and automatic secondary character animation in the movie and video game industries. Recent works in reinforcement learning have proposed impressive results for single character simulation, especially the ones that use imitation learning based techniques. However, imitating multiple characters interactions and motions requires to also model their interactions. In this paper, we propose a novel Multi-Agent Generative Adversarial Imitation Learning based approach that generalizes the idea of motion imitation for one character to deal with both the interaction and the motions of the multiple physics-based characters. Two unstructured datasets are given as inputs: 1) a single-actor dataset containing motions of a single actor performing a set of motions linked to a specific application, and 2) an interaction dataset containing a few examples of interactions between multiple actors.
Based on these datasets, our system trains control policies allowing each character to imitate the interactive skills associated with each actor, while preserving the intrinsic style. This approach has been tested on two different fighting styles, boxing and full-body martial art, to demonstrate the ability of the method to imitate different styles.
\end{abstract}

\begin{teaserfigure}
  \includegraphics[width=\textwidth]{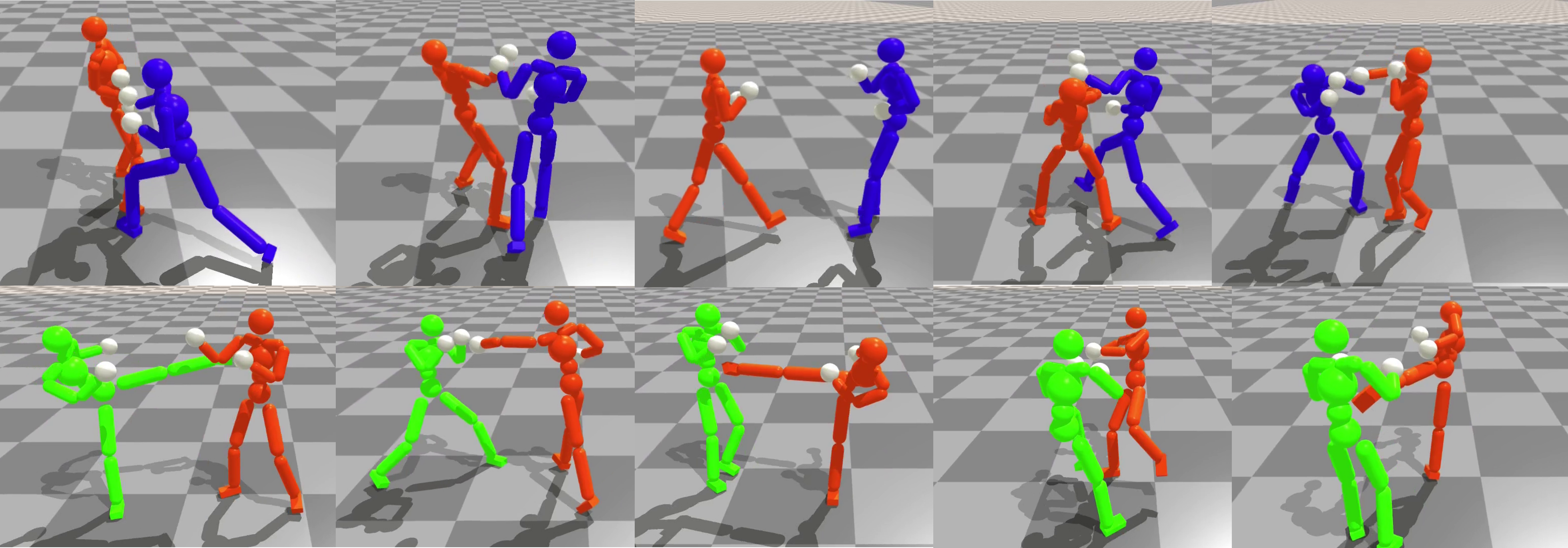}
  \caption{Two examples of imitation-based simulation with two different styles. Top: boxing scenario trained with single-actor and multiple-actors motion capture boxing datasets. Bottom: Qwankido (martial art) scenario based on the same approach, simply replacing the boxing datasets by Qwankido datasets. In these examples, the sequences show the ability of the characters to avoid attacks and then to counter-attack.}
  \Description{}
  \label{fig:teaser}
\end{teaserfigure}

\maketitle

\section{Introduction}
\label{intro}
Physics-based character control is an active field of research, as it enables to generate physically valid animations in complex interactive environments. In applications with multiple characters (simulated, or real-time avatars of users), these physics-based characters have to realistically interact with others in a large variety of situations.
In Virtual Reality fighting training, for example,
the user is not expecting an optimal behavior for the virtual opponent, but may prefer to prepare fighting against 
an opponent that imitates the behavior and the motions of a particular real boxer.
Such simulation should take the current state of the interaction into account, select the most relevant action to perform 
and compute the physically-valid corresponding motion, as a specific human would do, by imitating a small set of examples. 

Multiple physics-based character interactions can be simulated using spacetime constraints and optimal control \cite{Liu2006, Vaillant2017}. These approaches can find an optimal solution given a set of manually edited constraints, but may fail to imitate the style given in a small set of examples. 
Data-driven approaches select the optimal actions available in a database of examples, using game tree based methods \cite{Shum2008, Shum2012}.  
However, due to the simplicity of the rules and the high computational complexity, the intelligence of rule-based simulated characters is too limited to handle stylized interactions~\cite{Li2021}. 

Reinforcement learning has been explored for designing physics-based controllers capable of imitating motions given a small unstructured database of examples while achieving different tasks~\cite{peng2021amp}. This Adversarial Motion Priors (AMP) approach is based on the Generative Adversarial Imitation Learning (GAIL)~\cite{ho2016generative} framework.
It mainly uses an adversarial discriminator output as a reward instead of manually designing an imitation reward. However, it has only been applied to single character control. 
In this paper, we propose Multi-Agents Adversarial Interaction Priors (MAAIP), a method for imitating interactions and motions of multiple physics-based characters from unstructured motion clips. Our method is based on the Multi-Agent Generative Adversarial Imitation Learning (MAGAIL)~\cite{NEURIPS2018_240c945b} framework, and aims at extending AMP to deal with both the interaction and the motion of the controlled physics-based characters. Two unstructured datasets are used by the system: 1) a single-actor dataset containing motions of single actors performing a set of motions linked to a specific application, and 2) an interaction dataset containing few examples of interactions between multiple actors. 
Our system trains control policies allowing each character to imitate the interactive skills associated with each actor from the demonstrations, while preserving the intrinsic style. Similarly to AMP, the single-actor dataset is used to train a single motion prior, while 
the interaction dataset offers a novel complementary interaction prior to train each agent on how to behave in different interactive situations, with other agents.  
The interaction prior is therefore acting as a measure of similarity between the motions of the characters when interacting with each other, and the interaction examples in the datasets. The single motion prior offers a complementary repertoire of individual possible motions that may not appear, or not sufficiently, in the multiple-actors dataset. 
To the best of our knowledge, 
MAAIP is one of the first adversarial learning system 
for physics-based multiple-character animation that combines adversarial motion prior and interaction prior, allowing different characters to imitate interaction from a set of unstructured motion clips performed by multiple actors.

We evaluate our method by simulating competitive interactions between two physics-based characters, with different styles: boxing (hands only) and a Qwankido (a Sino-Vietnamese martial art with full body interactions). We use a few minutes of single-actor motion clip examples, and short sequences of interaction motion clips. We show the ability of our method to simulate interactions between two fighters, while imitating the style of each fighter contained in the datasets, without the need of designing specific constraints or rewards. We then explore the limits of the generalization ability of the method, when dealing with situations that have not been captured in the interaction dataset used for training. 
We also show how to control the interaction, by simply adding new rewards, such as interactively controlling the direction of the simulated fight, making the fighter be more aggressive, or more defensive.

\section{Related works}
\label{related}
Physics-based simulation relies on the dynamic equation of motion to generate joint angles trajectories for a character. However, the main challenge with these methods is to design a controller that generates realistic motions, with a desired style, and given a set of goals to achieve. In the two next sections, we review relevant physics-based simulation methods for a single (section~\ref{rel:physics}) and multiple (section~\ref{rel:multiple}) characters. We then introduce Imitation Learning techniques used for physics-based character simulation in section~~\ref{rel:imitation}.

\subsection{Single physics-based character control}
\label{rel:physics}
Physics-based character simulation has a long history in computer animation. Early efforts focused on developing locomotion control using motion analysis and hand-crafted controllers~\cite{hodgins1995animating},
abstract models~\cite{coros2010generalized}, optimal control~\cite{muico2011composite}, model predictive control~\cite{hamalainen2015online, mordatch2010robust} and reinforcement learning~\cite{xie2020allsteps, yin2021discovering}. These approaches typically require prior knowledge and hand-tuned parameters, which can make them difficult to apply to complex motions and scenarios. To address these difficulties,
several physics-based controllers have been supplemented with the motion capture data, using trajectory tracking to follow motion clips and a balance controller to keep the character upright \cite{zordan2002motion}. More recent works tracked reference motions by learning policies that get feedback from the physics simulation~\cite{lee2010data,liu2016guided}. With the development of deep reinforcement learning techniques, it became possible to robustly track agile human motions~\cite{peng2018deepmimic} and to generalize to various morphologies~\cite{won2019learning} and environments~\cite{xie2020allsteps}. 

\subsection{Multiple characters animation}
\label{rel:multiple}
Simulation of multiple characters interacting with each other 
involves defining properly the interaction between characters: 
relative positions between body parts of the characters~\cite{Ho2010}, but also more complex parameters, such as gaze, orientations or time coordination. 
Optimal control with spacetime optimization has been used to solve complex interaction problems involving multiple physics-based characters~\cite{Liu2006, Vaillant2017}. It requires careful design and tuning of the cost functions to obtain realistic simulations.
Other works proposed an offline game tree expansion to explore all the possible interactions between characters to simulate multiple characters competing or collaborating in a given scenario~\cite{Shum2008, Shum2012}.
All these approaches have been designed to find an optimal solution, but cannot easily imitate realistic behaviors contained in example motion capture clips.

When a few examples of interactions are available, reinforcement learning is a promising way to control physics-based characters. \cite{haworth2020deep} proposed a hierarchical policy that incorporates navigation, footstep planning, and bipedal walking skills, for controlling navigation of pedestrians. Unlike previous approaches, this method learns control policies that can handle interactions between multiple simulated humanoids. 
\cite{won2021control} proposed a two-steps approach that first learns an imitation policy from single-actor motion capture data, then transfers it into competitive policies. 
~\cite{liu2022motor} trained football teams of physically simulated humanoids in a sequence of training stages using a combination of imitation learning, single/multi-agent reinforcement learning and population-based methods.
However, these approaches have not been designed to leverage available interaction data of a few examples.
\subsection{Imitation Learning for physics-based character simulation}
\label{rel:imitation}
Imitation learning in physics-based animation uses reference motion data to improve the quality of the simulated motions. This is typically done by implementing a tracking objective, where the goal is to minimize the error between the simulated poses and example poses. 
This can be achieved through the use of a phase variable provided as an additional input to the controller for synchronization, or by providing target poses from the reference motion as inputs to the controller.~\cite{liu2010sampling, da2008simulation, kwon2017momentum, lee2010data, liu2016guided, peng2018deepmimic}.
However, using a single phase variable may not allow scaling to datasets containing multiple disparate motions, and using a reference pose as a target for the controller requires a high level controller to select the motions to imitate from as well as the manual definition of the pose error metrics~\cite{peng2017deeploco}.

Adversarial imitation learning~\cite{ho2016generative, ziebart2008maximum} is an alternative approach to avoid manually designing and tuning specific pose error metrics. It showed promising results to imitate motions, given a database of examples~\cite{merel2017learning, wang2017robust, xu2021gan}. This approach relies on an adversarial discriminator, aiming to distinguish  simulated motions from those depicted in the demonstration data. The discriminator is then used as a reward function to train a control policy to imitate the type of motion observed in the demonstration data. However, adversarial learning algorithms can be unstable during training, and the quality of the resulting motion can still be low compared to tracking-based methods. Adversarial Motion Priors (AMP)~\cite{peng2021amp} proposed a number of tweaks to address those issues, such as using gradient penalty, but did not handle interaction imitation of multiple characters.

Multi-Agent Generative Adversarial Imitation Learning (MAGAIL) \cite{NEURIPS2018_240c945b} is a promising framework to design controllers 
that imitate the interaction behavior of multiple characters, given a small set of unstructured motion capture examples.  
However, this framework has never been developed to control multiple physics-based characters. This raises the question of how to define the state representation to model the interaction of multiple characters, and how to build a discriminator based on interactions instead of just the motion of a single character. 

\section{System Overview}
\label{overview}
To produce realistic motions and interaction behaviors between multiple characters, we use two main databases (also denoted demonstrations):
\begin{itemize}
    \item a Multiple-actors motion capture dataset $\mathbb{M}^I$ that includes interaction between multiple actors. For our application, we use a dataset of fighting motions between two fighters of two different styles: Boxing (only the upper-body attacks) and QwanKiDo (full-body movements) 
    \item a Single-actor
    motion capture dataset $\mathbb{M}^S$ that includes basic skills of the same activity. 
    It enables simulated physics-based characters to have access to a larger repertoire of realistic motions than those included in the Multiple-actors dataset.
\end{itemize}

\begin{figure}[t]
    \centering
    \includegraphics[width=\columnwidth]{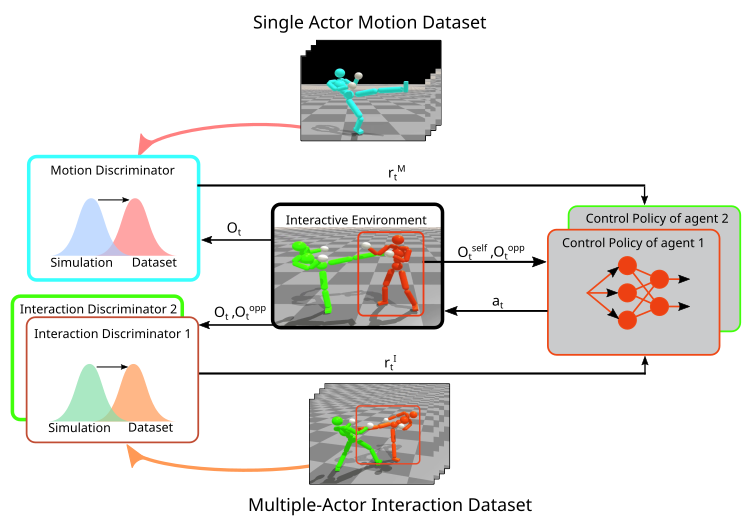}
    \caption{Overview of the system. Multiple-actors motion capture clips are used to train an interaction discriminator assigned to each agent, aiming at returning an interaction reward $r^I$. Single-actor motion clips are used to train a motion discriminator that returns a motion reward $r^M$. The rewards learned by the two discriminators are combined to train each agent's policy in order to imitate the interactive behavior depicted in the datasets.}
    \label{fig:overview}
\end{figure}

Figure~\ref{fig:overview} illustrates the overview of our approach. Each dataset $\mathbb{M}^I$ and $\mathbb{M}^S$ contains motion clips $\{m_S^{i} \in \mathbb{M}^S\}$ and $\{m_I^{i} \in \mathbb{M}^I \}$. The goal of the method is to simulate interaction behaviors and motions that imitate the style contained in the Multiple-actors interaction dataset $\mathbb{M}^I$. The Single Actor dataset $\mathbb{M}^S$ is used to 1) offer a wide variety of possible motions to the simulated characters, and 2) make the physics controller be more robust.  
Each motion clip can be seen as a sequence of character poses $m_{S}^{i}=\{q_{t}^{i}\}$ for the motion dataset $\mathbb{M}^S$, and as a sequence of two interacting characters poses $m_{I}^{i}=\{q_{t}^{i,0}, q_{t}^{i,1}\}$ for the interaction dataset $\mathbb{M}^I$, with two fighters denoted $0$ and $1$ respectively. Based on these poses $m_{I}^{i}$, we propose to build an observation at time $t$, $o_t$=[$o^{self}_t$, $o^{opp}_t$], for each character ($self$ for the agent, and $opp$ for the opponent). In section~\ref{model}, we give more details about the agent's observations.

We define the controller for each character using a policy: 
\begin{equation}
\pi (a_t |o^{self}_t , o^{opp}_t) \end{equation}
where $a_t$ is the action that specifies the set of target joint angles (target poses) used by a Proportional Derivative (PD)  
controller~\cite{tan2011stable}. Based on the physical model, the contact forces are computed during the simulation, both during the training and simulation phases. Thus, they can be used to simulate impacts, or design specific rewards minimizing self-damages or maximizing damages on the opponent.  

An adversarial discriminator is trained to compute a reward $r^{I}([o^{self}_t, o^{opp}_t], o^{self}_{t+1})$. 
For each character, this discriminator is trained to distinguish between interactions simulated by the simulated agent from those shown in the demonstrations (Multiple-actors motion capture dataset). Hence, it is possible to train specific discriminators for each character, with his specific style. This is a key idea here, as we expect to be able to generate individual style for each character in the final multiple-characters simulation.  

The observation's transition $(o^{self}_t , o^{self}_{t+1})$ is also used to compute a motion reward $r^{M}(o^{self}_{t}, o^{self}_{t+1})$ that measures the naturalness of the simulated motion. Similarly, $r^{M}$ is the output of an adversarial discriminator trained to differentiate between generated motions and demonstrations stored in the Single-actor motion capture dataset. 

The two learned rewards could be combined with other rewards $r^{C}_t$, to offer control facilities, such as maximizing physical contact on a specific body part of the opponent, or driving the interaction to a given direction.

\section{Method}
\label{method}
We formulate the interaction imitation problem as a Partially Observable Markov Game, where the goal is to learn optimal policies of multiple agents interacting with each other in the same environment~\cite{littman1994markov,busoniu2008comprehensive}. In the following, we detail our architecture adapted from Multi-Agent Generative Adversarial Imitation Learning~\cite{NEURIPS2018_240c945b}, with two major contributions: modeling the interaction with an opponent, and new objectives for training the system.

\subsection{Multi-Agent Generative Adversarial Imitation Learning}
\label{MAGAIL}
Multi-Agent Generative Adversarial Imitation Learning (MAGAIL) \cite{NEURIPS2018_240c945b} is a variant of the Generative Adversarial Imitation Learning (GAIL)~\cite{ho2016generative} that is used to deal with multi-agent interactions. In MAGAIL, multiple agents $i$ (each with their own policy $\pi_{\theta_{i}}$) are trained to imitate the behavior of one or many expert policies $\pi_{E_{i}}$, using a Generative Adversarial Network framework~\cite{goodfellow2020generative}.
For each agent $i$, a parametrized discriminator $D_{\omega_i}$ maps state action-pairs $(s_t, a_t)_i$ to scores that are optimized to discriminate expert demonstrations generated by unknown expert policy $\pi_{E_{i}}$ from behaviors produced by the agent's policy $\pi_{\theta_{i}}$. $D_{\omega_i}$ plays the role of a reward function for the generator $\pi_{\theta_{i}}$, which in turn attempts to train the agent to maximize its reward, therefore fooling the discriminator~\cite{NEURIPS2018_240c945b}. The objective to be optimized is the following:
\begin{equation}
    \min_{\theta} \max_{\omega} \mathbb{E}_{\pi_{E}}[\sum_{i=1}^N \log D_{\omega_i}(s, a_i)] + \mathbb{E}_{\pi_{\theta}}[\sum_{i=1}^N \log(1 - D_{\omega_i}(s, a_i))]
\end{equation}

where $\pi_{\theta}$ denotes the joint policy for $N$ agents $\pi_{\theta} = \prod_{i=1}^{N} \pi_{\theta_i}$ and $\pi_{E}= \prod_{i=1}^{N} \pi_{E_i}$ denotes the joint policy for $N$ experts. The policies $\pi_{\theta_i}$ are updated through reinforcement learning 
by using as a reward function for each agent $i$:
\begin{equation}
    r_t^i = -\log(1-D_{\omega_i}(s_t, a_t)_i)
\end{equation}

\subsection{Self and opponent observations}
\label{model}
The observation of each agent $o_t$ = [$o^{self}_{t}$, $o^{opp}_{t}$] consists of a set of features describing the proprioceptive configuration of its own body $o^{self}_{t}$ at the current time $t$, as well as features describing the current observation about the opponent $o^{opp}_{t}$. 
The features used to model $o^{self}_{t}$ are:
\begin{itemize}[leftmargin=*]
    \item Root's height from the ground $\in \mathbb{R}$
    \item All body parts' positions in the character's local coordinate frame $\in \mathbb{R}^{42}$
    \item All body parts' local rotations $\in \mathbb{R}^{90}$
    \item All body parts' local linear and angular velocities $\in \mathbb{R}^{45}$
\end{itemize}
We used a reduced set of features for observations about the opponent compared to the one used in~\cite{won2021control}. Each agent's features about the opponent $o^{opp}_{t}$ include:
\begin{itemize}[leftmargin=*]
    \item Opponent's root position $\in \mathbb{R}^{3}$, orientation $\in \mathbb{R}^{6}$, linear and angular velocities $\in \mathbb{R}^{3}$ in the current character's local coordinate frame
    \item Opponent's head, torso, hands and feet positions and velocities in the current character's local coordinate frame $\in \mathbb{R}^{18}$
\end{itemize}
We use the linear and angular velocities as relevant information for deciding the appropriate reaction to the opponent. In the context of physical interaction between two characters, we assume that the controller should benefit from potential anticipation skills thanks to this type of information. Indeed, in real competitive or collaborative interactions between people, this anticipation skill is important.
Similarly to previous works~\cite{peng2022ase, peng2021amp}, the pelvis segment is assumed to be the root of the character. The local coordinates are then expressed in this reference frame, with the x-axis oriented along the root facing direction, and the y is up. The body parts' rotations are encoded using two 3D vectors corresponding to the tangent and normal of its link 
local coordinate frame, expressed in the link parent’s coordinate frame. 
The observation space obtained from these features has a dimension of 274. 
The actions $a_t$ correspond to target poses used by the PD controller to compute joint torques for the character's joints. 
The target pose for spherical joints is represented by 3D exponential map $q \in \mathbb{R}^3$~\cite{grassia1998practical} such that the rotation axis $v$ is computed by $v = \frac{q}{||q||_2}$ and the rotation angle $\theta = ||q||_2$. This representation is more compact than 4D axis-angle or quaternion representations, and also avoids the gimbal lock issue in Euler angles~\cite{peng2021amp}.
The target rotations for revolute joints are specified as 1D rotation angles $q = \theta$.
The resulting action space has 28 dimensions. 
\subsection{Adversarial Motion and Interaction Priors}
\label{priors}
In order to imitate close interaction from motion capture demonstrations, we use a learned reward function $r^{M}$ that takes into account the motions generated by each simulated character $i$. We also use a learned interaction reward $r^{I}$ that takes into account its behavior with respect to the opponent. We use a combination of these two rewards to train each agent with RL:
\begin{equation}
    r(o_t, a_t, o_{t+1}) = w^{M}r^{M}(o^{self}_{t}, o^{self}_{t+1}) + w^{I}r^{I}(o_{t}, o^{self}_{t+1})
    \label{eq:rewards_combination}
\end{equation}
where $w^{M}$ and $w^{I}$ are weights associated with the two rewards functions $r^{M}$ and $r^{I}$ respectively. 

Following~\cite{peng2021amp}, the single motion prior $D^{M}$ is modeled by a learned discriminator trained to predict whether an observation transition $(o^{self}_{t}, o^{self}_{t+1})$ is a real sample from the dataset, or a sample simulated by the agent. 
We model the interaction reward by learned discriminators, each one assigned to an agent. Given the interaction dataset $\mathbb{M}^I$
of multiple actors, each discriminator $D^I$ is trained to predict if the transition $(o_{t}, o^{self}_{t+1})$, i.e. the reaction of the agent with respect to the other one, is within the distribution of the demonstrations. 

Since we use demonstrations from unlabeled and unstructured motion capture clips, we do not have access to actions needed by MAGAIL, as introduced in section~\ref{MAGAIL}. Therefore, we train the motion discriminator $D^M$ with the observation transitions $(o^{self}_{t}, o^{self}_{t+1})$, and the interaction discriminators $D^I$ with transitions $(o_{t}, o^{self}_{t+1})$ as inputs, as suggested in previous works~\cite{torabi2018generative}. In this case, the reward function based on the motion discriminator is given by:
\begin{equation}
    r_{t}^{M} = -\log(1-D^{M}(o^{self}_{t}, o^{self}_{t+1}))
    \label{eqn:motion_reward}
\end{equation}
while the reward based on the interaction discriminators is:
\begin{equation}
    r_{t}^{I} = -\log(1-D^{I}(o_{t}, o^{self}_{t+1}))
    \label{eqn:interaction_reward}
\end{equation}
We also use the gradient penalty regularization~\cite{peng2021amp} in order to stabilize the training of the discriminators and improve the quality of generated behaviors. Therefore, with $\phi=(o^{self}_{t}, o^{self}_{t+1})$, the objective for training the single motion prior $D^{M}$ is formulated by:
\begin{equation}
\begin{split}
    \min_{D^{M}}-\mathbb{E}_{\pi_{E}}[\log D^{M}(\phi)] - \mathbb{E}_{\pi_i}[\log (1-D^{M}(\phi))] \\
    + w_{gp} \mathbb{E}_{\pi_{E}}\left[\left|\left|\nabla_{\phi} D^{M}(\phi)|_{\phi}\right|\right|^{2}\right] 
\end{split}
\label{eqn:motion_disc_objective}
\end{equation}
where $\pi_E$ denotes an unknown expert policy that generated the demonstration transitions, $w_{gp}$ is a manually specified coefficient. On the other hand, with $\psi=(o_{t}, o^{self}_{t+1})$, the objective for training each interaction prior $D^{I}$ is:
\begin{equation}
    \begin{split}
    \min_{D^{I}}-\mathbb{E}_{\pi_{E}}[\log D^{I}(\psi)] - \mathbb{E}_{\pi_i}[\log (1-D^{I}(\psi))] \\
    + w_{gp} \mathbb{E}{\pi_{E}}\left[\left|\left|\nabla_{\psi} D^{I}(\psi)|_{\psi}\right|\right|^{2}\right] 
\end{split}
\label{eqn:interaction_disc_objective}
\end{equation}

\subsection{Network Architecture}
\label{architecture}
Since the agents are homogeneous (i.e. they have the same observation and action spaces), we use parameter sharing for their policies, so that all agents share the same network. Previous works have shown that this makes the learning be more efficient~\cite{yu2021surprising,terry2020revisiting,christianos2021scaling}. Therefore, the policies $\pi$ are modeled by a neural network for which the inputs are the full observation $o_t$ of each agent $i$ as well as an indicator of the identity of the agent $i$, and outputs the mean $\mu(o_t, $i$)$ of a Gaussian distribution over actions $\pi(a_t|o_t, $i$)=N(\mu(o_t, $i$);\Sigma)$ where the covariance matrix $\Sigma$ is fixed during training. It is a fully-connected network consisting of 3 hidden layers of 1024, 1024, 512 units with ReLU activations~\cite{nair2010rectified}, followed by a linear output layer. We also use centralized training and decentralized execution (CTDE) for training the agents~\cite{lowe2017multi}. Therefore, we use a centralized value function $V(s_t=(o^{0}_t,o^{1}_t))$ shared by the two agents during training that takes as input the concatenation of all agents’ local observations to build a global state $s_t$ ~\cite{lowe2017multi}. The value function $V(s_t)$, the interaction discriminators $D^{I}$ and motion discriminator $D^{M}$, are modeled as networks with similar architecture.

\subsection{Training}
We use the framework of MAGAIL~\cite{NEURIPS2018_240c945b} with the multi-agent proximal policy optimization algorithm MAPPO~\cite{yu2021surprising, schulman2017proximal}: at each time step $t$, each agent receives a local observation $o_t = [o^{self}_{t}, o^{opp}_{t}]$ from the environment and decides an action $a_t$. Then, it receives an interaction reward $r_t^{I}$ and a motion reward $r_t^{M}$, computed from their respective discriminators, and possibly a control reward  $r_t^{C}$ specified by the user, to add a level of control to the interaction of the characters. Similar to~\cite{peng2021amp}, we use a combination of these rewards to get the final imitation reward $r_t$ at time $t$ according to Equation~(\ref{eq:rewards_combination}).
To stabilize the training in tasks where additional control rewards are used, we use reward scheduling so that at the beginning of the training, agents learn first to imitate motions from the single motion datasets then we introduce later the rewards for interaction and then the control reward. We find that by using this strategy, the resulting interaction is more convincing and does not collapse to unwanted behavior because of opposing rewards.

After collecting a batch of trajectories with the policies, we record them in buffers to update the policy networks, the centralized value function $V$, and the discriminators $D^I$ and $D^M$, similarly to~\cite{peng2021amp}. We also add replay buffers $B^{I}_i$ for each interaction discriminator $D^{I}$ associated with each agent $i$.

We use Generalized Advantage Estimation GAE($\lambda$)~\cite{schulman2015high} to compute advantages for updating the policies. The centralized value function is updated using TD($\lambda$)~\cite{sutton1988learning}. We follow the recommendations from~\cite{yu2021surprising} to choose the hyperparameters of the multi-agent PPO algorithm. The training process is described in algorithm~\ref{algo:maaip_algorithm}.

\begin{algorithm}[]
\caption{Training Algorithm for Multi-Agent Interaction policies}
\label{algo:maaip_algorithm}
\begin{algorithmic}[1]
\REQUIRE Initialized policies $\pi$, interaction discriminators $D^{I}$, motion discriminator $D^{M}$ and value function $V$ 
; Single-Actor motion dataset $\mathbb{M}^S$; Multi-Actor interaction dataset $\mathbb{M}^I$
\ENSURE Learned policies $\pi$ and reward functions $D^{I}$ and $D^{M}$
\WHILE{learning is not done}
\STATE $\mathbb{B}^{\pi}$, $\mathbb{B}^{M}$, $\mathbb{B}^{I}$ $\leftarrow$ $\emptyset$ initialize data buffers for each agent.
\FOR{trajectory $k=1,...,m$ of length $T$}
\STATE $\tau^k$ $\leftarrow$ {$(o_t, a_t)^{T-1}_{t=0}$} collect trajectory rolled out with policies $\pi$ for all agents
\FOR{timestep $t=0, ..., T-1$}
\STATE $d_{t}^{M}$ $\leftarrow$ $D^{M}(o^{self}_{t}, o^{self}_{t+1})$ get score from the single motion prior for all agents
\STATE $d_{t}^{I}$ $\leftarrow$ $D^{I}(o_{t}, o^{self}_{t+1})$ get scores from interaction priors for all agents
\STATE $r_{t}^{M}$ $\leftarrow$ calculate motion reward according to formula 4.
for all agents
\STATE $r_{t}^{I}$ $\leftarrow$ calculate interaction reward according to formula 5.
for all agents
\STATE $r_{t}$ $\leftarrow$ combine $r_{t}^{M}$ and $r_{t}^{I}$ according to formula 2.
\STATE record $r_{t}$ in the trajectory $\tau^k$ for each agent.
\STATE store transitions $(o^{self}_{t}, o^{self}_{t+1})$ in $\mathbb{B}^{M}$ for all agents.
\STATE store transitions $(o_{t}, o^{self}_{t+1})$ in $\mathbb{B}^{I}$ for each agent.
\ENDFOR
\STATE store trajectory $\tau^k$ in $\mathbb{B}^{\pi}$ for each agent.
\ENDFOR
\FOR {update steps $i = 1, . . . , n$}
\STATE update $D^{M}$ using K transitions sampled from $\mathbb{M}^S$ and from $\mathbb{B}^{M}$ according to formula 6.
\STATE update each $D^{I}$ using K transitions sampled from $\mathbb{M}^I$ and from $\mathbb{B}^{I}$ according to formula 7.
\ENDFOR
\STATE update $\pi$ and $V$ using samples from $\mathbb{B}^{\pi}$ for all agents using MAPPO.
\ENDWHILE
\end{algorithmic}
\end{algorithm}

\section{Experiments and results}
\label{results}
We carried out experiments on two scenarios: boxing, where the agents only used displacements and upper-body actions, and QwanKiDo, a Sino-Vietnamese martial art involving full-body actions.
 
We first evaluated the standard case, using the imitation reward only~(\ref{eq:rewards_combination}), in an application where the two characters had to imitate interactions of the demonstrations. Then, we showed that adding a task-specific reward for minimizing (resp. maximizing) the damage received by (resp. given to) each character leaded to simulate more defensive (resp. aggressive) behaviors. We also demonstrated an example of controlling the moving direction while keeping the interaction. Finally, we pushed the system to the limits by simulating interaction between characters that were trained on different sets of demonstrations.

\subsection{Experimental setup}
\label{motion_capture}
The unstructured dataset used for training agents on fighting interactions contains motions of two different fighting styles: boxing and QwanKiDo.  
We used a Qualisys opto-electronic motion capture system, composed of 22 Oqus 200Hz cameras, to track 46 anatomical landmarks placed according to the Qualisys animation marker set guidelines. When contact occurred, some markers may fly away, so that the corresponding samples were eliminated.  The data were downsampled to 30Hz and retargeted to the character's skeleton used in the simulation. Some examples of motion capture sessions in boxing and QwanKiDo are given in the supplementary video. We plan to share our datasets upon acceptance of this work to support future research in filling the gap between individual motor skills for single characters and interactive skills between multiple characters.

Isaac Gym~\cite{makoviychuk2021isaac} was used for the physics-based simulation engine for GPU-based accelerated training. We simulated 2048 environments in parallel on a single NVIDIA A6000 GPU, each with 2 agents. We ran the simulations at a frequency of 60Hz with 2 sub-steps, while the policies were queried at 30Hz. All policies were trained for 2 billion steps, which takes approximately 15 hours of training time. The algorithm's hyperparameters are available in supplementary material.

\vspace{0.1cm}
\textbf{Boxing Scenario.}
\label{boxing}
The boxing scenario involves two characters who can displace and use their upper-body to attack (with jabs, crosses, hooks and uppercuts),  or defend (using guard, slipping, swaying, parrying, blocking and clinching). For the Single-actor motion dataset, 4 high-level volunteer boxers (1 professional and 3 regional-level competitors) participated in a single full-body motion capture session. The resulting single-boxer dataset contained approximately 15 minutes of boxing. 
For the Multiple-characters dataset, we asked pairs of the above boxers to perform 30s to 90s rounds. For each trial, the opponents started far away from each other, to capture some displacement toward a real opponent. Two pairs of boxers participated in this experiment, with different personalized "specials" (considered as styles). The total duration of multiple-actors dataset was 3 minutes.

\vspace{0.1cm}
\textbf{QwanKiDo scenario.}
\label{qwankido}
The QwanKiDo scenario also involves two characters, but the repertoire of possible motions is larger, including kicks, elbow or knee strikes, and sweeping. The protocol was similar to the one used for boxing, with 2 participants, single actor and two-actors sessions. The total usable motion capture duration for the single-actor dataset was 10 minutes. This scenario raises more challenges for the imitation approach, as the quantity of available demonstrations is smaller, whereas the number of possible actions is larger. Moreover, the "specials" for each fighter are visually more different than those observed for the boxing scenario. The total duration for the multiple-actors dataset was 3 minutes. 

\subsection{Fighting simulation using the priors only}
\label{res:dynamicauto}
In this first application, we only used the rewards computed from the discriminators' outputs.
We used the weighting values of $w^{M}=0.2$ for the motion reward and $w^{I}=0.8$ for the interaction reward in Equation~(\ref{eq:rewards_combination}).
For the interaction, each agent was associated with the same opponent in all the demonstrations, assuming that it should enable to provide this specific opponent style of interaction to this agent. 

\begin{figure}
  \centering
    \subfloat{\includegraphics[scale=0.15]{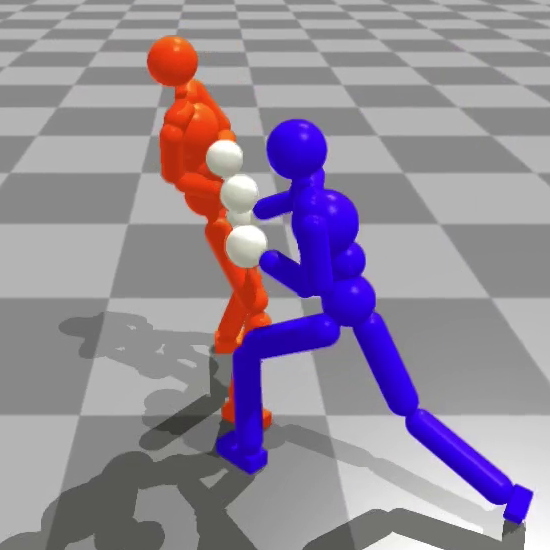}}\hfill
    \subfloat{\includegraphics[scale=0.15]{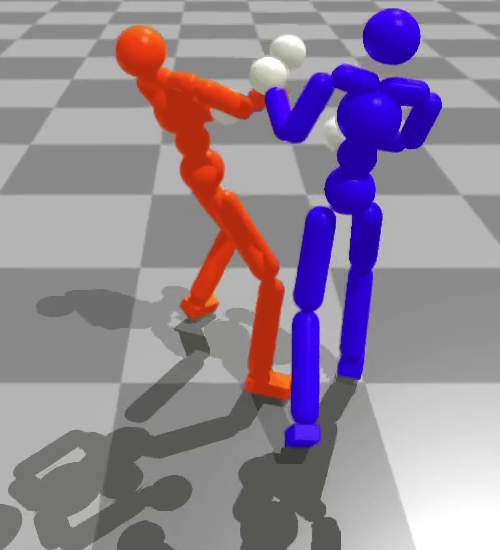}}\hfill
    \subfloat{\includegraphics[scale=0.15]{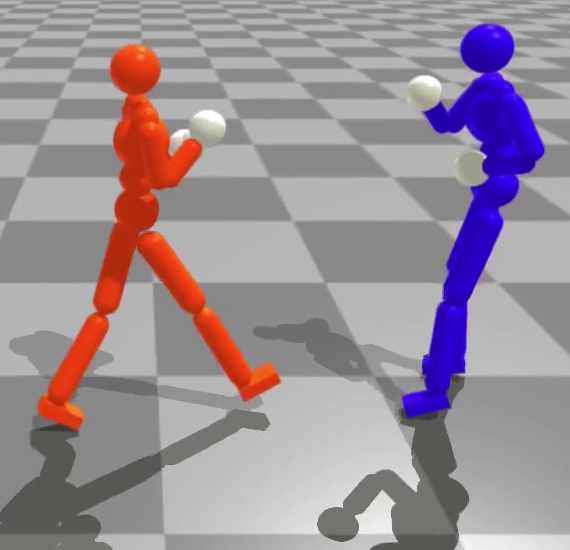}}\hfill
    \subfloat{\includegraphics[scale=0.15]{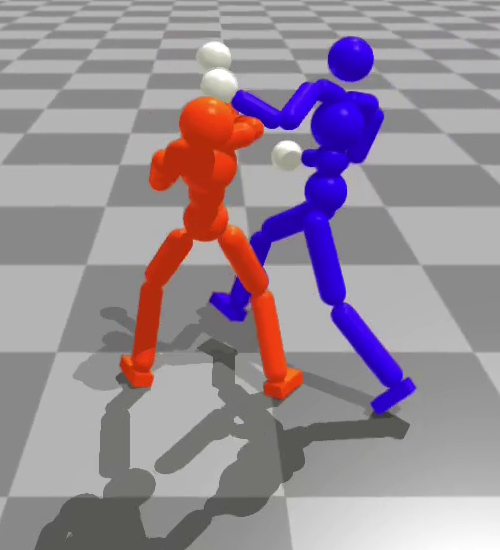}}\hfill
    \subfloat{\includegraphics[scale=0.15]{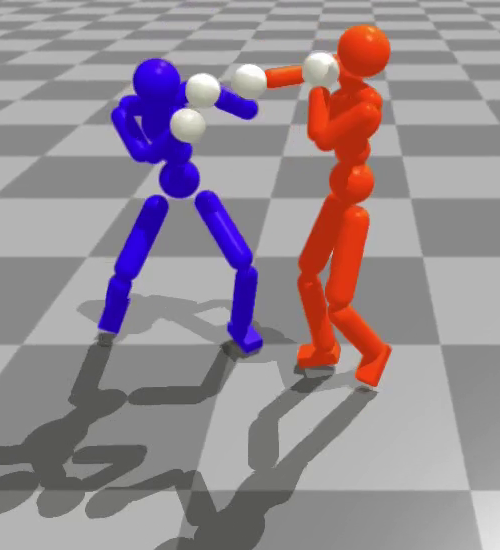}}\hfill
    \subfloat{\includegraphics[scale=0.15]{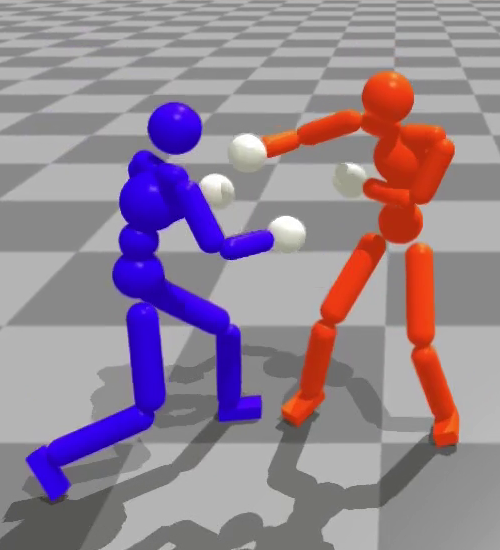}}
\caption{Simulation of boxing interaction between two agents. The boxers show agility in the movements, interactive skills such as getting closer to the opponent, dodging and blocking attacks as well as finding attack openings.}
\label{fig:boxing_interaction}
\end{figure}

\begin{figure}
  \centering
    \subfloat{\includegraphics[scale=0.12]{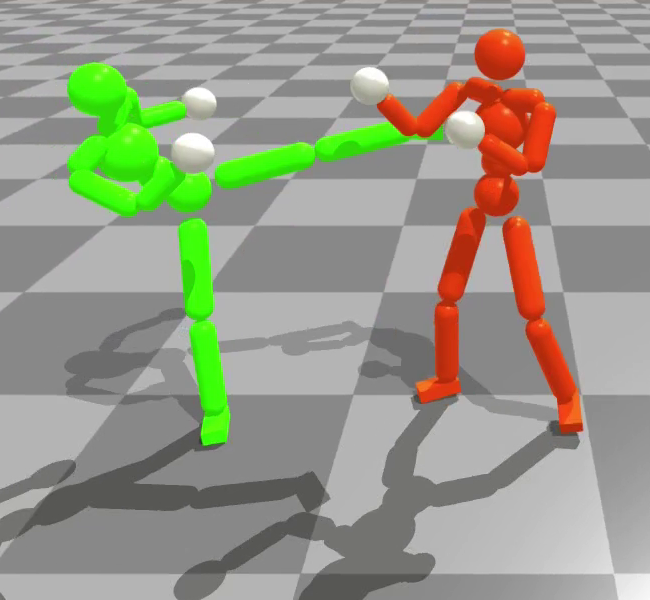}}\hfill
    \subfloat{\includegraphics[scale=0.12]{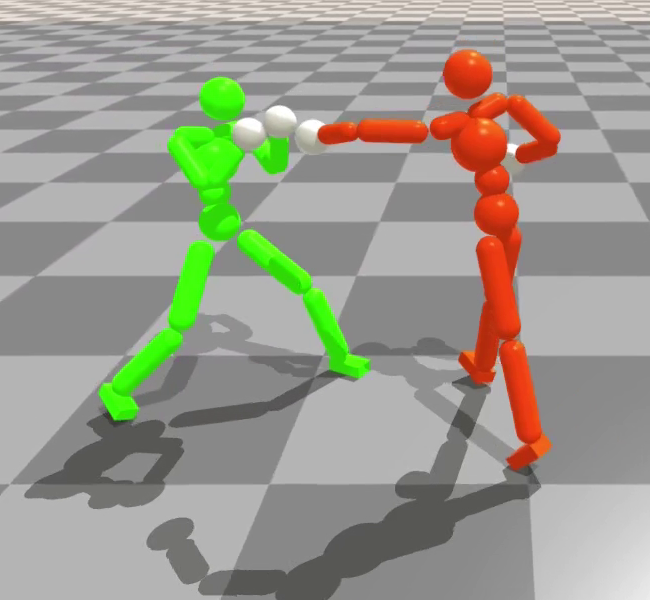}}\hfill
    \subfloat{\includegraphics[scale=0.12]{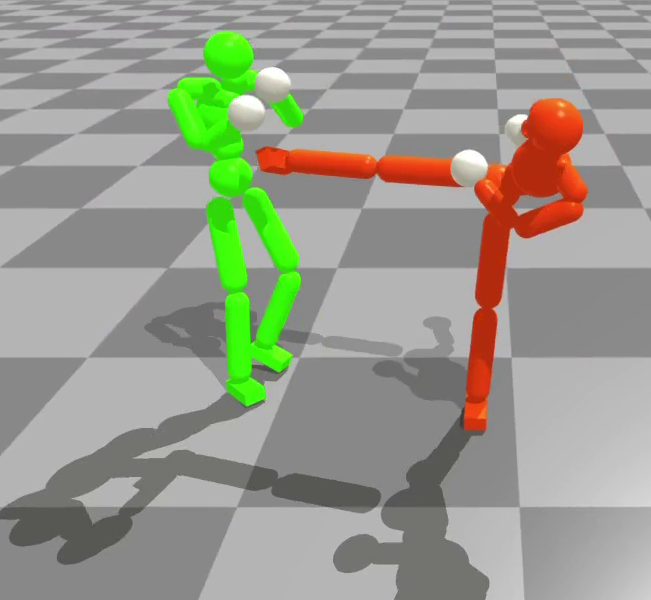}}\hfill
    \subfloat{\includegraphics[scale=0.12]{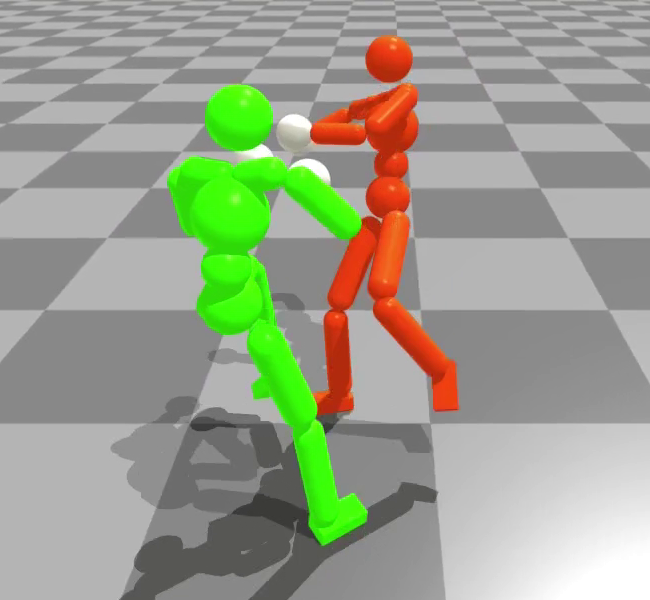}}\hfill
    \subfloat{\includegraphics[scale=0.12]{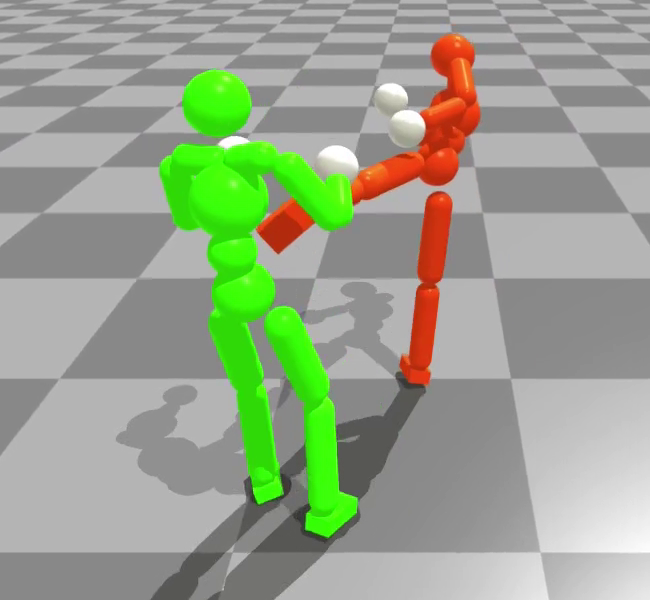}}\hfill
    \subfloat{\includegraphics[scale=0.12]{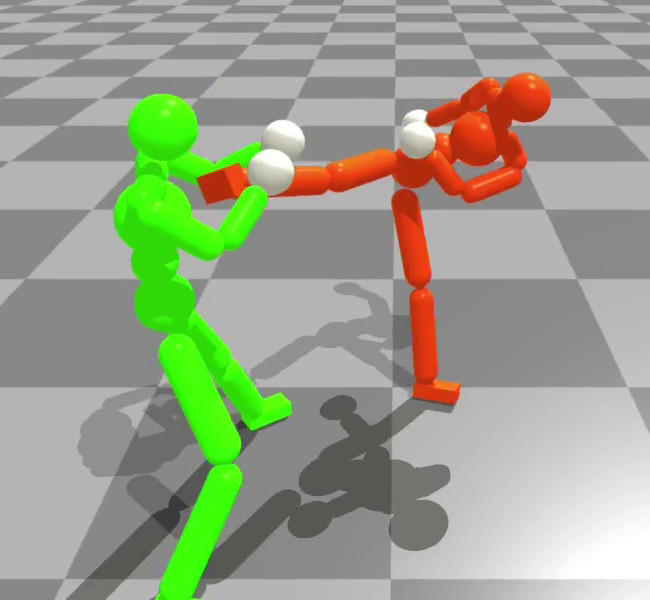}}
\caption{Simulation of QwanKiDo interaction between two agents.  The agents show highly-dynamic motions, such as using full body for attacks, unique fighting styles similar to the actor motions used for training them.}
\label{fig:qwankido_interaction}
\end{figure}

Visual results are depicted in Figures~\ref{fig:boxing_interaction} and \ref{fig:qwankido_interaction}. In the resulting sequences, one can see that the fighters learned basic fighting skills, such as getting closer to the opponent, staying in guard stance when approaching, anticipating openings for attacks and evading incoming attacks. They also learned footwork skills for fighting as they move around the opponent and remain at a safe distance before switching to attack. The experts who participated in the motion capture sessions were able to recognize the participant who served as a demonstration for each avatar, in all the simulations. This result should of course be confirmed by a scientific perceptual study. 

We ran numerous simulations, with random initialization states (global position and orientation) and obtained very convincing results, as shown in the supplementary video. In very few cases, we could obtain clearly unrealistic results, which demonstrates one of the fundamental limits of imitation-based approaches: too few examples in the demonstrations may lead to simulate unrealistic behaviors. These unrealistic behaviors could be strange following behaviors, or repeating the same motion many times (due to mode collapse of the discriminators). To partly mitigate this risk, the system could be trained with more examples, and could also use additional rewards, such as minimizing or maximizing impacts, which should provide a wider set of potential solutions.

\subsection{Fighting simulation using additional task-dependent control rewards}
\label{res:dynamicdamage}
To control the generated interaction and guide the selection of the motions the agents should imitate from the dataset, we tested additional task rewards $r^{C}$. Firstly, we introduced a reward that encourages the agents to minimize the damage dealt by the opponent to specific body parts. Secondly, we designed another reward that encourages maximizing damages on the opponent. These task-specific rewards are reasonable choices for both boxing and QwanKiDo. 

The additional rewards could enable the system to find acceptable solutions when facing a new situation that was not captured in the single motion and interaction priors. Compared to previous works, the new behaviors are generated automatically, without the need of designing a specific motion planner for motion selection. 

\begin{figure}
  \centering
    \subfloat{\includegraphics[scale=0.15]{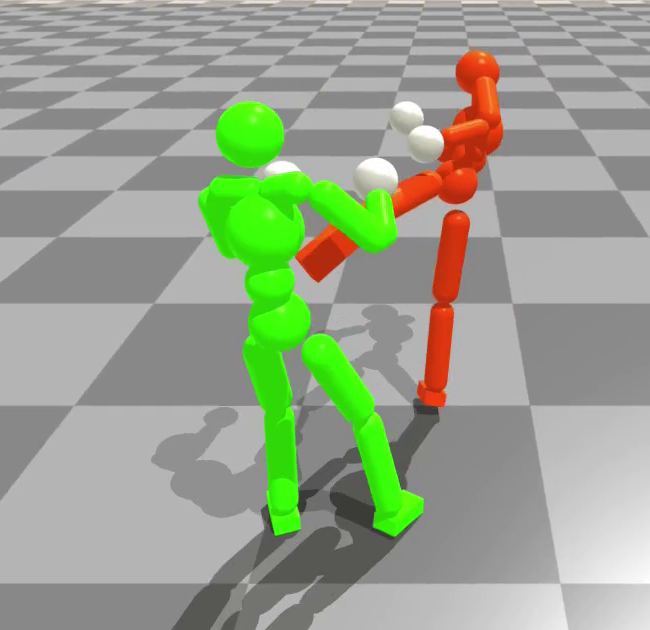}}\hfill
    \subfloat{\includegraphics[scale=0.15]{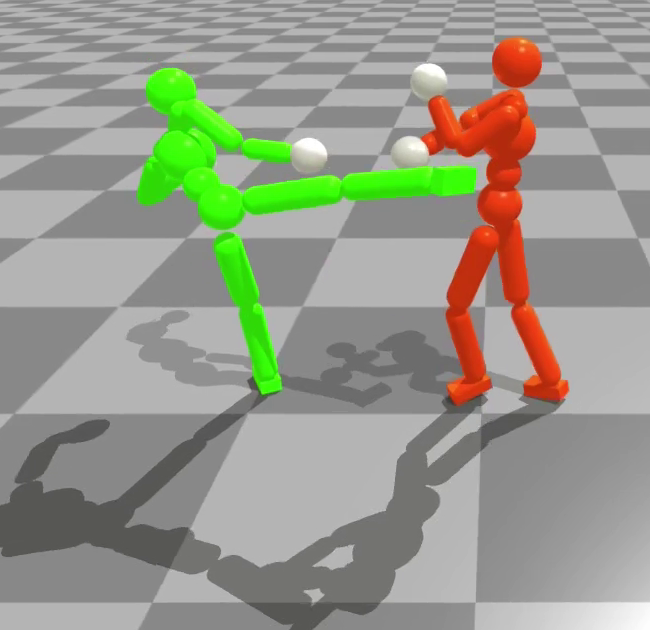}}\hfill
    \subfloat{\includegraphics[scale=0.15]{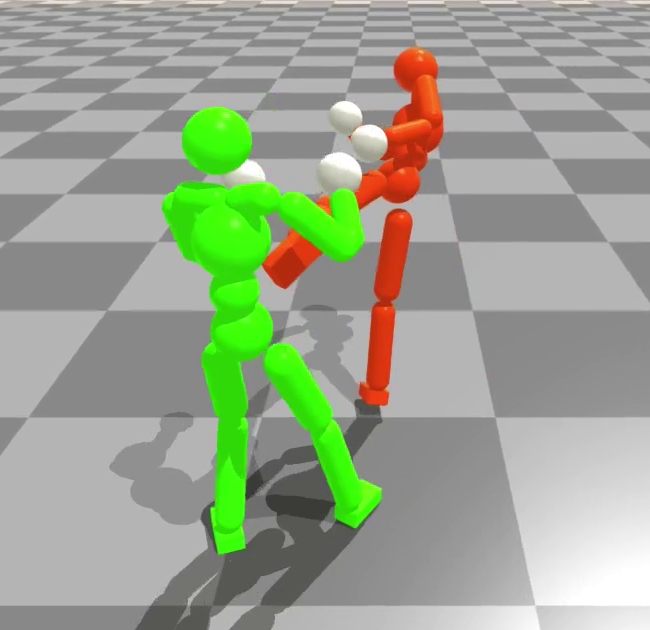}}\hfill
    \subfloat{\includegraphics[scale=0.15]{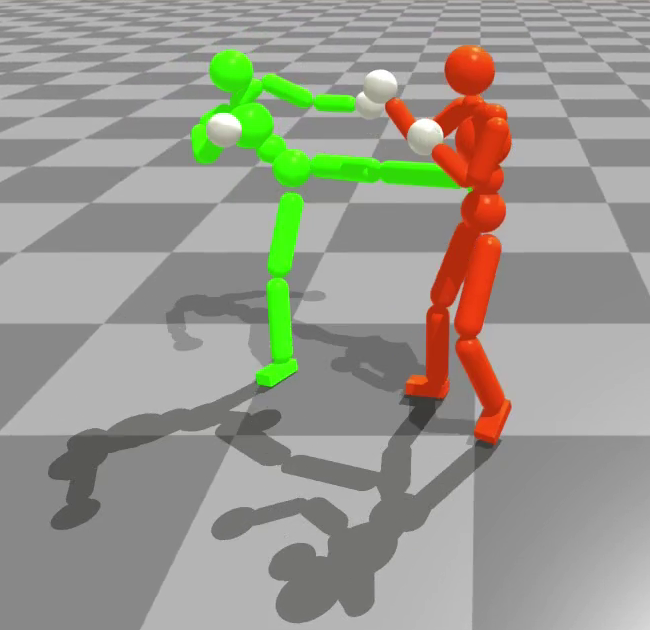}}\hfill
    \subfloat{\includegraphics[scale=0.15]{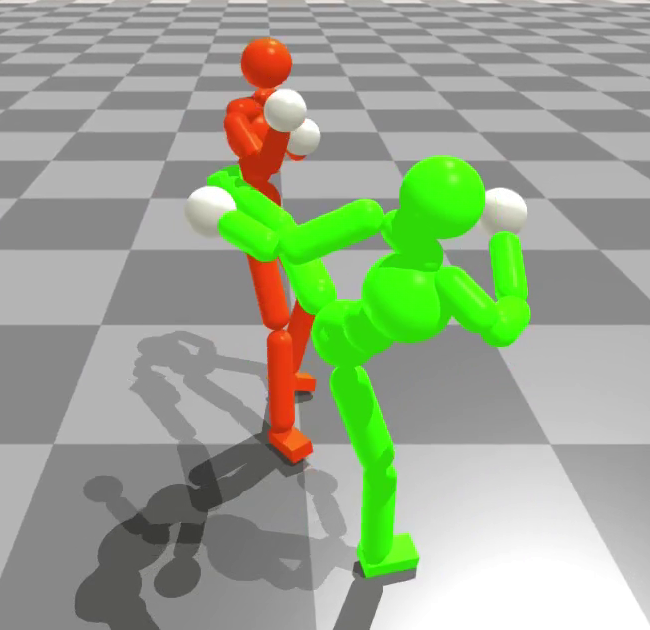}}\\
    \subfloat{\includegraphics[scale=0.15]{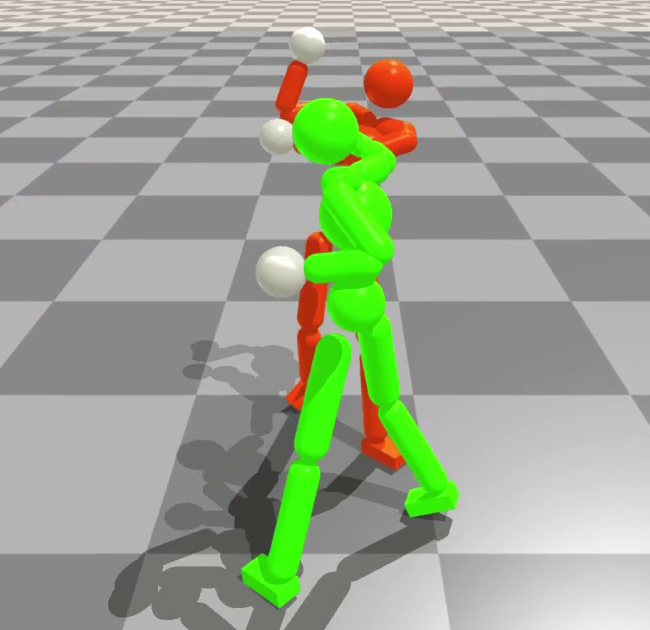}}\hfill
    \subfloat{\includegraphics[scale=0.15]{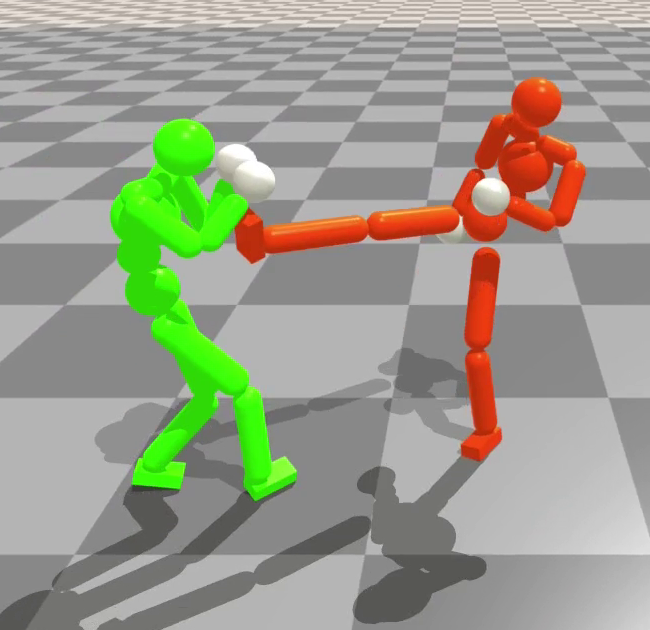}}\hfill
    \subfloat{\includegraphics[scale=0.15]{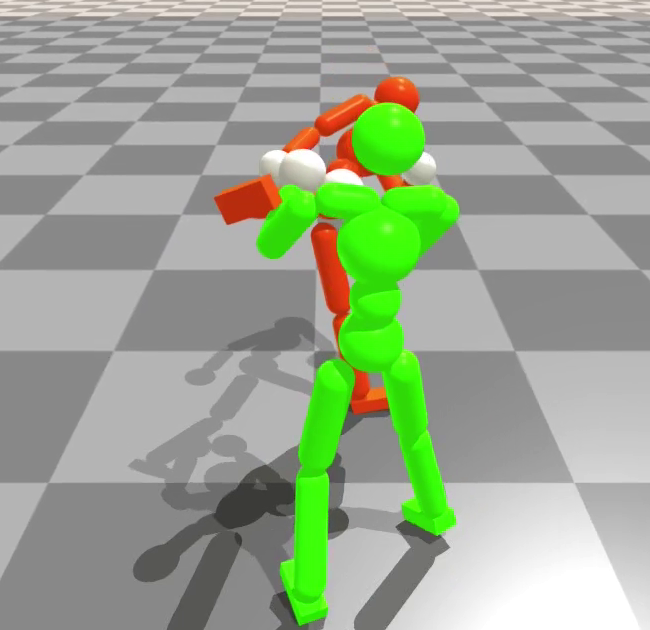}}\hfill
    \subfloat{\includegraphics[scale=0.15]{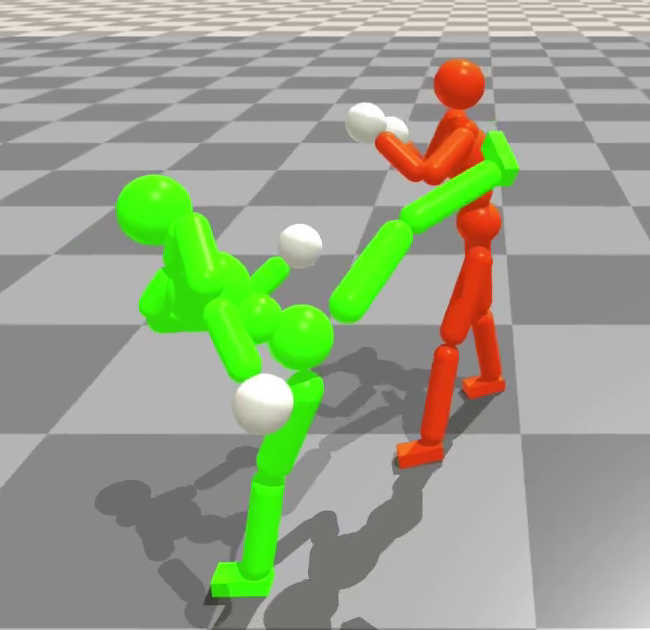}}\hfill
    \subfloat{\includegraphics[scale=0.15]{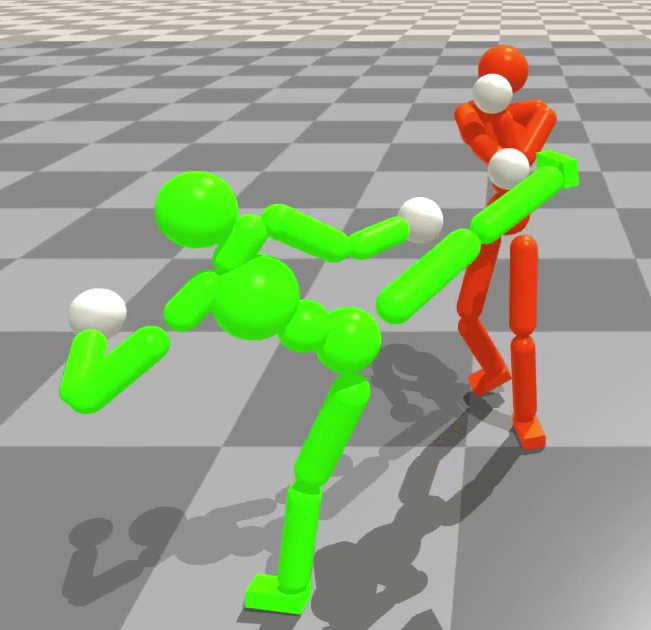}}
\caption{An example of fighting simulation using additional control rewards, that encourages the agents to minimize the damage dealt by the opponent to specific body parts: head, torso and pelvis. Top: without control reward ; Bottom: with control reward. The reward drives the agents into simulating interactive behavior where they act more defensively, and they block attacks more often.}
\label{fig:interaction_damage_reward}
\end{figure}

Let $|f_{opp \rightarrow self}|$ be the magnitude of external contact normal forces applied by an opponent (considered as "damages") to the head, torso and pelvis of a character. 
The damage minimization reward is then given by:
\begin{equation}
    r^{C} = exp({-w \cdot |f_{opp \rightarrow self}|}).
\end{equation}
Similarly, the damage maximization reward is expressed by:
\begin{equation}
    r^{C} = 1 - exp({-w \cdot |f_{self \rightarrow opp}|})
\end{equation}
The weighting for the different rewards becomes: $w^{M}=0.1$, $w^{I}=0.4$ and $w^{C}=0.5$. 
We computed the "damages" applied to each character by averaging the total damage received over 32 trials, with an episode length of 1200 frames, with and without using these task rewards.
The quantitative results (see Table~\ref{tab:damage_table}) showed a significant decrease of the "damages" with the damage minimization reward compared to using the imitation reward only. Reversely, we noticed an increase in the received damage when using the damage maximization reward.
\begin{table}[htb]
    \centering
\begin{tabular}{|l|cc|cc|cc|}
\hline
Scenario & \multicolumn{2}{l|}{Imitation only}    & \multicolumn{2}{l|}{Damage min.} & \multicolumn{2}{l|}{Damage max.} \\ \hline
Boxing Duo 1                   & \multicolumn{1}{l|}{2210} & 3261 & \multicolumn{1}{l|}{820}   & 862   & \multicolumn{1}{l|}{6759}  & 6143  \\ \hline
Boxing Duo 2                   & \multicolumn{1}{l|}{1135} & 2010 & \multicolumn{1}{l|}{957}   & 1393  & \multicolumn{1}{l|}{9861}  & 8146  \\ \hline
Qwankido                      & \multicolumn{1}{l|}{4038} & 2216 & \multicolumn{1}{l|}{123}    & 215   & \multicolumn{1}{l|}{8623}  & 9435  \\ \hline
\end{tabular}
    \caption{Mean damage values (in Newton) for 32 randomly initialized episodes of length 1200 each, with imitation reward only, minimizing or maximizing damage additional reward. The damages are cumulative contact forces applied to the head, the torso and the pelvis, either of the controlled character (to minimize damages) or of the opponent (to maximize damages). 
    }
    \label{tab:damage_table}
\end{table}
The top part of Figure~\ref{fig:interaction_damage_reward} depicts a QwanKiDo sequence simulated without the damage minimization reward, leading to a series of attacks. The bottom part of Figure~\ref{fig:interaction_damage_reward} depicts the resulting sequence when adding the damage minimization reward, which shows more defensive and less engaging behavior. 

\subsection{Target Heading Task}
\label{res:control_two}
\begin{figure}
  \centering
    \subfloat{\includegraphics[scale=0.18]{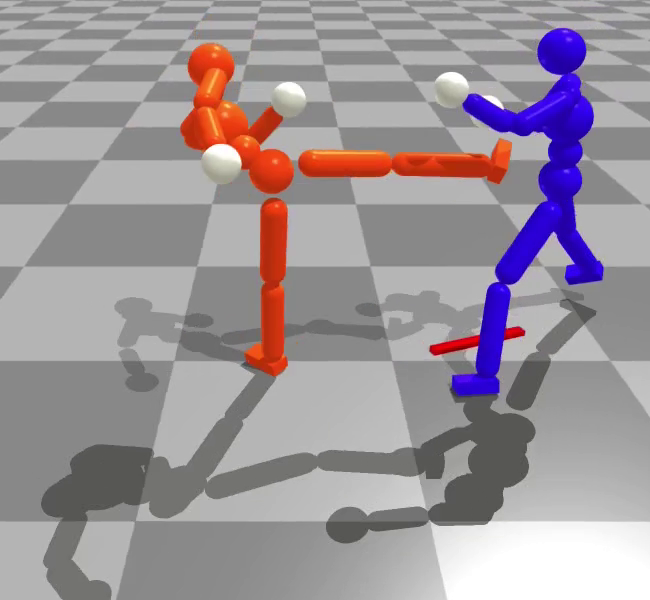}}\hfill
    \subfloat{\includegraphics[scale=0.18]{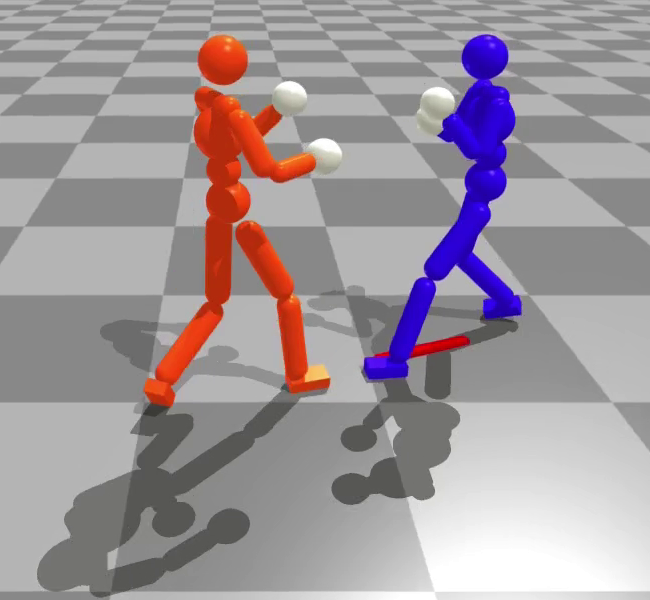}}\hfill
    \subfloat{\includegraphics[scale=0.18]{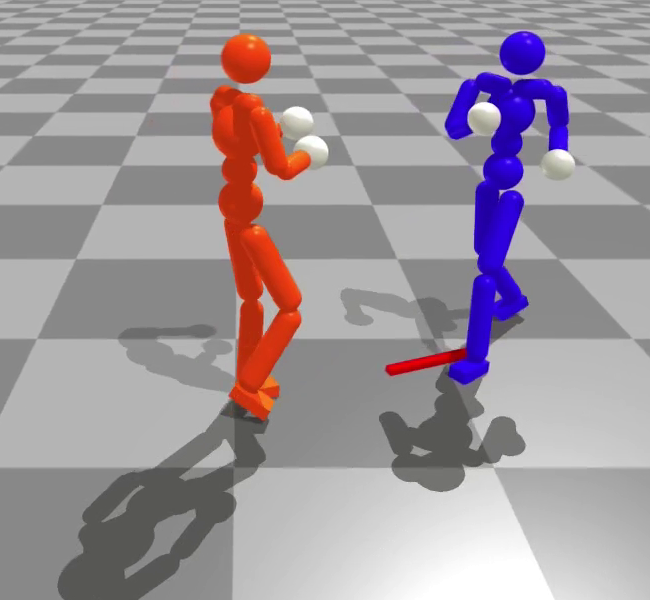}}\hfill
    \subfloat{\includegraphics[scale=0.18]{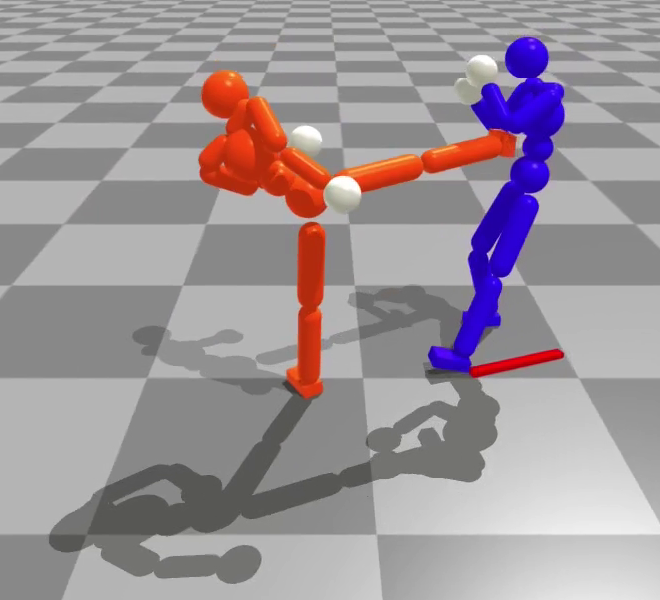}}\\
    \subfloat{\includegraphics[scale=0.18]{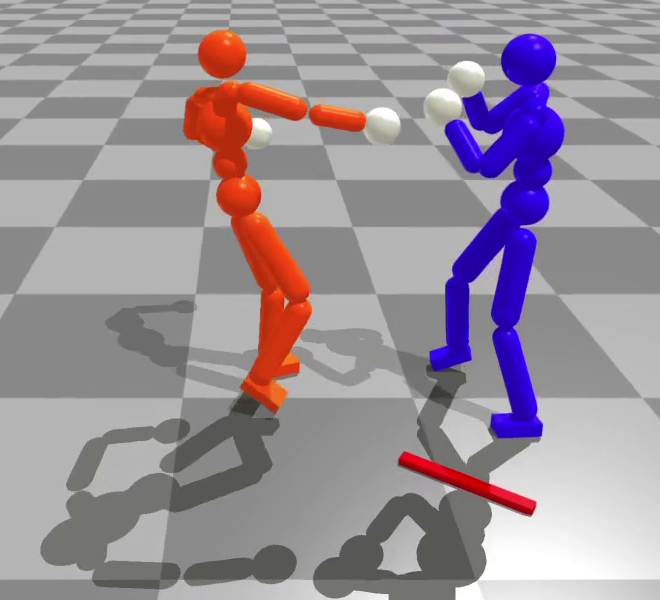}}\hfill
    \subfloat{\includegraphics[scale=0.18]{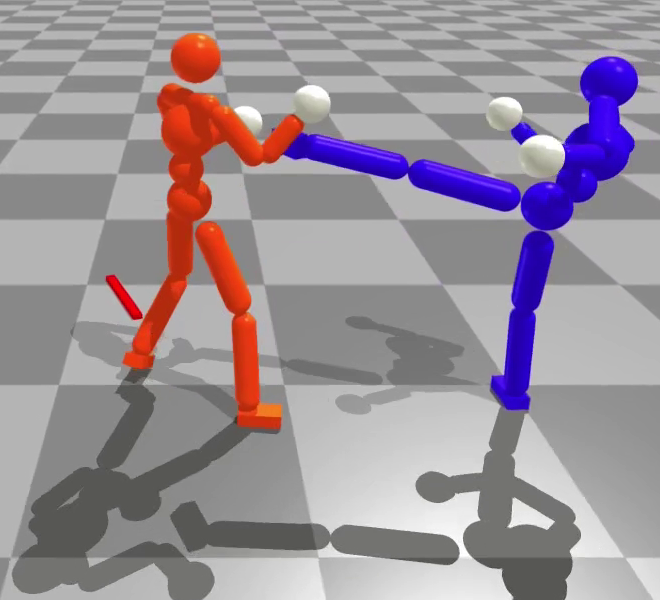}}\hfill
    \subfloat{\includegraphics[scale=0.18]{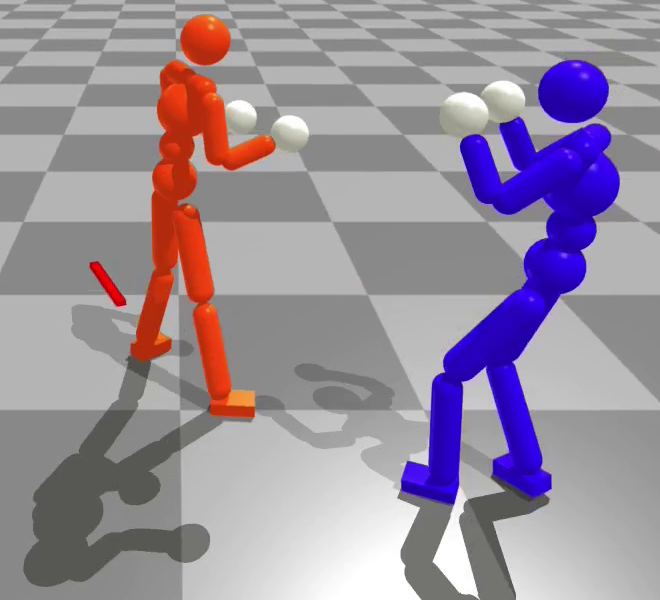}}\hfill
    \subfloat{\includegraphics[scale=0.18]{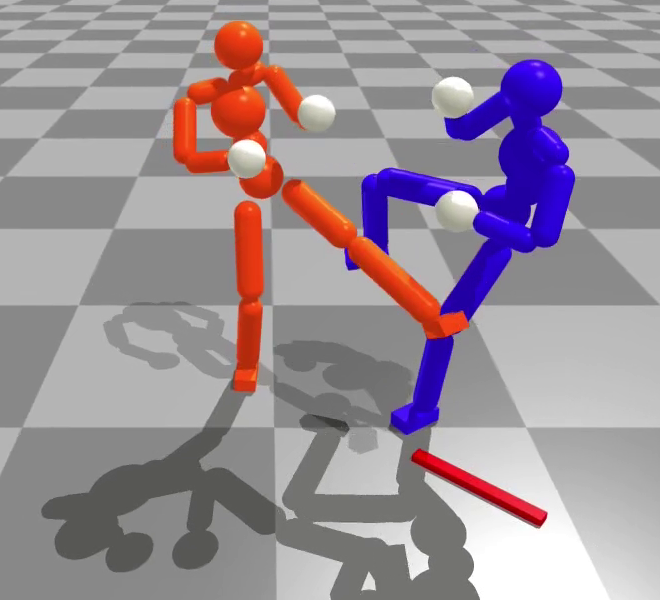}}
\caption{An example of interaction simulation with heading controls. The fighters are constrained to move towards a given target direction, represented by the red line.}
\label{fig:heading_task_interaction}
\end{figure}

In this task, the objective for the characters is to move along an imposed target heading direction $d^*$, while still fighting one against each other.
We conditioned the policies of the agents on the given target direction in the local coordinate frame for each character
$d^{*}_{t}$ at time t, and we used a reward function similar to the one used in~\cite{peng2021amp}:
\begin{equation}
    r^{C} = exp({-w \cdot ( d^* \cdot v^{root}  )})
\end{equation}
where $v^{root}$ is the root velocity for each character. 
The weighting used for this task is $w^{M}=0.1$, $w^{I}=0.4$ and $w^{C}=0.5$. Figure~\ref{fig:heading_task_interaction} shows the interaction of two QwanKiDo fighters moving towards a given direction. The resulting task return of the heading control task for QwanKiDo and Boxing is reported in Table~\ref{tab:heading_control_table}. 
The resulting animation is shown in the supplementary video.

This task in particular illustrates the interest of the single motion prior.
The results show that characters trained with the single motion prior slightly better follow the heading direction, with slightly better task return $r^C$. Although the agents trained without the single motion prior might obtain a good task return, they only can imitate the motions included in the interaction dataset, which can lead to unnatural behavior. Indeed, some selected displacements may exhibit hits or avoidance to satisfy the heading constraints, while these actions are not appropriate in the current situation: avoidance without opponent attack, or punches while the opponent is too far. This type of artifacts was not observed when also using the single motion prior. 

Let us notice that the training for this task is very sensitive to the weights associated with each component. Indeed, when giving more importance to the single motion prior with a high $w^{M}$ weight, the simulated agents follow the given direction without interacting with each other, as some displacement without interaction are available in the single motion prior. Reversely, when giving more importance to the interaction prior $w^{I}$, the agents mainly use displacements based on hits and avoidance, as interaction-free displacements are rare in the interaction prior (see the supplementary video for some examples).

\begin{table}[htb]
    \centering
\begin{tabular}{|l|c|c|}
\hline
Scenario & With Single MP & Without Single MP \\ \hline
Boxing Duo1                   & 0.86                             & 0.82                                \\ \hline
Boxing Duo2                   & 0.80                             & 0.76                                \\ \hline
QwanKiDo                      & 0.90                             & 0.89                                \\ \hline
\end{tabular}
    \caption{Performance of the trained agents in the heading control task when using or not the single motion prior (MP). The performance is quantified by the average normalized task return $r^{C}$ for 32 episodes of 500 length each. 
    }
    \label{tab:heading_control_table}
\end{table}
\subsection{Transfer to unseen fighting situations}
\label{res:unseen}
In our approach, the main idea is to imitate an interaction given in a multiple-characters motion capture dataset. 
To evaluate if our method could handle novel and unseen fighting situations, we trained two agents with different interaction datasets. Let us consider for example that the agent $0$ is trained with the boxing dataset, and another agent $1$ with the QwanKiDo dataset. The agent $1$ had seen some examples of attacks performed with the arms, such as jabs or uppercuts, although they may have been performed with a different style. However, agent $0$ had never seen any kick or sweeping attacks. Again, it is hard to quantify the ability of the system to generalize, as there are no real metrics to quantify the realism of the resulting simulation. 

We found that the agents were able to keep the basic interactive skills, such as getting closer to the opponent, facing him and staying on guard, even though they were not trained against those specific opponents. However, we also noticed that they performed fewer attacks and are less engaging, as attacks are conditioned by a given observation of the opponent, and there is no such an attack signal for observations that have never been seen during training. We believe that enhancing the datasets used for training and using policy architectures that account for past observations should help to handle a larger variety of fighting situations, but it may still suffer from distributional shift~\cite{ross2010efficient}.

\section{Ablation Study}
In this section, we study the importance of the components of our method by ablating the sensitivity to the weighting between the interaction prior and the single motion prior, as well as the impact of the losses used for training the discriminators.

\subsection{Single Motion Prior Impact}
The single motion prior in our framework aims at providing single actor motion examples to generate natural behavior and account for unseen situations in the interaction prior. We show the importance of using it in the heading control task, introduced previously. In this task, we found that using only the interaction prior may lead to similar task returns (see Table~\ref{tab:heading_control_table}), but the resulting motions were less natural. Indeed, the agents seem to exploit the motion included in the interaction to achieve high reward at the cost of motion naturalness, especially when the given direction changes. As the single motion prior is trained with a larger variety of displacement motions compared to the interaction prior, it enables to generate more natural foot work and displacements. Therefore, it enables to create seamless transitions between interaction and displacement motions (see supplementary video). However, the weighting between the interaction prior, the single motion prior and the task reward needs to be tuned so that the agents achieve the desirable behavior as a high weighting for the single motion prior might lead to agents that completely ignore the interaction and focus on maximizing the heading task relying only on the single motion prior.

For the transfer to unseen fight situations introduced in~\ref{res:dynamicdamage}, we found that adding the single motion prior helps to generate behaviors in fighting situations which are not present in the interaction dataset, and yields more plausible results in general, compared to when using only the interaction prior. Indeed, the agents trained with only the interaction prior struggle more to keep natural behavior in out-of-distribution states. However, we found that the generated behavior is sensitive to the weighting assigned to the single motion prior. By assigning more importance to the motion prior, the characters are less interactive and focus more on maximizing the motion reward. Consequently, they start punching/kicking far from each other (see the example of such case in the supplementary video). We believe that better strategies for varying the weights assigned to each term depending on the task could be beneficial to improve the quality of the resulting interaction rather than having constant weights.

\subsection{Discriminators Training Loss Impact}
The objective used for training the single motion prior and each interaction prior in our framework is the same one defined in the original GAIL~\cite{ho2016generative} which uses a sigmoid cross-entropy loss function. This loss function is known for training instability because of saturation of the sigmoid function, leading to vanishing gradients. To counter this, the authors of AMP proposed to use the loss function for least-squares GAN (LSGAN)~\cite{mao2017least} that showed more training stability and better overall quality. The objective for optimizing the discriminator is defined as: 
\begin{equation}
    \min_{D^{M}}  \mathbb{E}_{\pi_{E}}\left[{\left(D^{M}(\phi) - 1\right)}^{2}\right] + \mathbb{E}_{\pi}\left[{\left(D^{M}(\phi) + 1\right)}^{2}\right]
\end{equation}
with $\phi=(o_{t}, o_{t+1})$. The policy $\pi$ is then optimized using the following reward function:
\begin{equation}
    r(\phi) = \max\left[0, u - v \cdot \left( D^{M}(\phi) - 1 \right)^2\right]
\end{equation}
$u$ and $v$ are offset and scale to bound the reward between $[0, 1]$.

We experimented with this objective for training both the single motion prior and the interaction priors in the imitation task. We found that the quality of the generated interactive behavior is worse compared to what we get with the standard GAIL objective, and that it is more prone to mode collapse by repeatedly generating the same subset of motion sequences. Although the agents were able to perform the basic fighting motions included in the single motion dataset, their interactive capabilities were limited even when assigning more importance to the interaction priors in the total reward. We think that this degradation in interaction quality is due to the difficulty of solving the least-squares regression by the interaction priors when the environment is non-stationary in the setting of multiple characters' interaction. We show examples of these behaviors in the supplementary video.

\section{DISCUSSION AND LIMITATIONS}
\label{conclusions}
We have introduced an innovative adversarial system designed to replicate the intricate fighting interactions between multiple physics-based characters, utilizing unstructured motion clips. Building upon the foundations of the MAGAIL framework, our approach incorporates crucial adaptations to effectively simulate multiple physics-based characters' behaviors. The first significant enhancement involves the modeling of reactive behavior, wherein we establish a transition from the current full observation, that includes the self-observation of the agent itself and the current observation about the opponent, to the subsequent self-observation. This transition captures the dynamic nature of the characters' responses to their opponent, resulting in a more authentic simulation. Additionally, we devised a training strategy that encompasses both single motion and interaction priors.
The resulting sequences do not simply imitate the reference motions with the same frame order, but exhibits similar interactive behaviors to the interaction dataset by maximizing the rewards assigned by each prior. Hence, our approach enabled us to imitate the personalized reaction of fighters with specific styles. We can also provide the users with some control of the simulation, by adding task-specific rewards: following a given direction, minimizing the received impacts or maximizing damages to the opponents when searching for the next action, while still imitating the style of the interaction dataset. We could imagine other rewards, such as aiming specific parts on the opponent's body. The results show that although the interaction dataset could be enough to learn motion and interaction imitation policies, associating a complementary single motion prior helps to generalize to a wider range of situations with realistic motions. This is a key contribution of this work. 

However, like other Generative Adversarial Networks (GAN)-based methods, our approach can suffer from mode collapse: repeating the same interaction behavior and generating only a small subset of the interactions contained in the demonstrations, especially because of the multi-modality of the interaction dataset. Recent work~\cite{juravsky2022padl} tried to mitigate this problem by conditioning the motion prior on latents that encode each motion clip. Some other work~\cite{nguyen2017dual,hardy2019md} propose to use multiple discriminators to handle the multi-modality of the training distribution. Although these methods introduce new challenges such as predefining the number of discriminators to be used, increasing the number of trained parameters or the assumption of having a labeled reference motion dataset, we believe that they can serve in reducing the effect of mode collapse and improving the quality of the generated behavior. For example, if the motion clips are segmented and labelled, we could imagine using a discriminator for attacks, another for defense, etc.

On the other hand, our method can also be used to simulate new individual styles, or new multi-characters activities (fencing, dancing, collaborative work, etc.), by retraining the same system but with new single-character and multiple-characters datasets. However, this can also be a limitation, as it requires providing enough examples to make the physics-based character correctly imitate the activity. Instead of fully retraining the policies,
it should be possible to use transfer learning: pretraining the system with basic skills, such as moving around while maintaining balance, and then fine-tuning the resulting policies with a few new specific examples. This is specifically true for simulating different individual styles for the same activity, where the basic actions should be very similar. 
For some activities, the effort required to capture interaction datasets of multiple actors would be an important obstacle. For applications in the movie industry, we could also imagine using animation sequences designed by animators to convey a specific style for imaginary characters.

While the motion generated by our framework is qualitatively similar to the motion of the motion clips examples, the resulting motion of some sequences may still appear unnatural; Since the method's goal is to imitate the style of the interactions given as examples, for safety reasons, it was difficult to ask the subjects to exert high impacts on the opponent, given that they were equipped with hard markers that could injure them. Hence, we asked them to perform shadow style combat with low impacts, which is actually imitated by the system. We have shown that the same framework works for (shadow) boxing and Qwankido by simply changing the input databases of examples, and in some cases the additional attack reward can lead to combat engagement that was not present in the original motions. We could expect that fighting motions with higher impacts would help to imitate real fights.

In this work, we only tested activities involving two fighters. Future investigations and tests are needed to check the capability of the system to scale to more characters and to adapt to different types of interaction such as dancing, where the choreography, synchronization and long duration contacts of multiple dancers are important for generating plausible results. We note that with the current policies' architecture, our system can only imitate short term reactions, such as parrying a strike, or counter-attacking with one strike. It cannot handle middle or long-term strategies involving a sequence of actions. We would like to explore techniques that incorporate high-level long term planning in the imitation learning process so that fighters are equipped with strategic play that they can learn from demonstrations while being able to use the same strategic reasoning in new fighting situations. Learning basic fighting skills with a low level controller, then learning strategic play from demonstrations by a high level controller equipped with a long term memory component would be an interesting future direction for this work.

\begin{acks}
This work was supported by French government funding managed by the National Research Agency under the Investments for the Future program in France 2030 with the grant ANR-18-EURE-0022 (DIGISPORT). It was granted access to the HPC resources of IDRIS under the allocation 20XX-AD011013491R made by GENCI.
\end{acks}

\newpage
\onecolumn

\newcommand{\htiwofig}{./fig/interaction_damage_reward/without_damage_reward}
\newcommand{\htiwfig}{./fig/interaction_damage_reward/with_damage_reward}

\newpage
\twocolumn
\bibliographystyle{ACM-Reference-Format}
\bibliography{paperbib}

\end{document}